%% file: main.tex
\documentclass{jdmdh}
\usepackage{array}
\usepackage{pgfplots}
\pgfplotsset{compat=1.18}

\usepackage{graphicx}
\usepackage{booktabs}
\usepackage{tikz}
\usepackage{adjustbox}
\usepackage{makecell}

\newcommand{\site}[0]{\url{https://tales.ko64eto.com}}
\usepackage{euflag}

\usepackage{enumitem}
\newlist{tabitemize}{itemize}{1}
\setlist[tabitemize]{nosep,
                  topsep= 0pt,
                  partopsep=0pt,
                  leftmargin= *,
                  label=\textbullet,
                  }

\input{tikz-header}

\titlespacing*{\section}
{0pt}{2.5ex plus 1ex minus .2ex}{1.3ex plus .2ex}
\titlespacing*{\subsection}
{0pt}{1.0ex plus 1ex minus .2ex}{1.3ex plus .2ex}
\titlespacing*{\subsubsection}
{0pt}{1.0ex plus 1ex minus .2ex}{1.3ex plus .2ex}

\title{Values That Are Explicitly Present in Fairy Tales: \\ Comparing Samples from German, Italian and Portuguese Traditions}
\author[1]{Alba {Morollon Diaz-Faes}}
\author[1,2]{Carla Sofia {Ribeiro Murteira}}
\author[3]{Martin Ruskov}
\affil[1]{NOVA University of Lisbon, Portugal} 
\affil[2]{School of Communication and Media Studies, IPL, Portugal
}
\affil[3]{University of Milan, Italy} 

\corrauthor{Martin Ruskov}{martin.ruskov@unimi.it}

\begin{document}

\maketitle

\abstract{Looking at how social values are represented in fairy tales can give insights about the variations in communication of values across cultures. We study how values are communicated in fairy tales from Portugal, Italy and Germany using a technique called word embedding with a compass to quantify vocabulary differences and commonalities. We study how these three national traditions differ in their explicit references to values. To do this, we specify a list of value-charged tokens, consider their word stems and analyse the distance between these in a bespoke pre-trained Word2Vec model. We triangulate and critically discuss the validity of the resulting hypotheses emerging from this quantitative model. Our claim is that this is a reusable and reproducible method for the study of the values explicitly referenced in historical corpora. Finally, our preliminary findings hint at a shared cultural understanding and the expression of values such as Benevolence, Conformity, and Universalism across the studied cultures, suggesting the potential existence of a pan-European cultural memory.}


\input{sections/010-intro}
\input{sections/020-literature}
\input{sections/030-method}

\input{sections/040-results}
\input{sections/050-conclusion}

\section*{Acknowledgments}

\euflag\ This work has received funding from the European Union's Horizon 2020 research and innovation programme under grant agreement No 101004949. This document reflects only the author's view and the European Commission is not responsible for any use that may be made of the information it contains.

\bibliographystyle{plainnat}
\bibliography{bibliography/acl2020,bibliography/corpora,bibliography/new,bibliography/theory}

\appendix\footnotesize

\section{Annex 1}

\label{sec:appendix}

The source code of the annotation analysis tool is available at:\\
\url{https://github.com/umilISLab/moreever/}.

On the pages below Table~\ref{tab:texts} lists the complete textual sources that were used in the corpus and Table~\ref{tab:tokens} the complete list of value-bearing tokens that were used.

\input{figures-tables/tab090-texts}



\input{figures-tables/tab096-tokens-reduced}

\end{document}

%% file: sections/010-intro.tex
\section{Introduction}\label{sec:intro}


Culture is defined “as a common heritage of a set of beliefs, norms, and values”~\citep{dhhs_us_mental_2001}, that influences an individual’s cognition and behaviour~\citep{wong_positive_2013}. Social values are understood as standards or criteria of the desirable, thus they guide the selection or evaluation of behaviours, policies, people, and events~\citep{schwartz_convergence_2020}.  Building on this understanding of values as a cornerstone of culture, we turn to literature as a mirror reflecting these values across different cultural contexts in the past. Developments in natural language processing (NLP), in particular word embeddings, have allowed for the quantitative analysis of historical corpora~\citep{miaschi_contextual_2020,rodriguez_word_2022}.

With this work we want to test the limits of an approach for studying the social values present in fairy tales, one of the most widely spread forms of popular narratives. Fairy tales are a privileged genre for the identification of patterns of cultural exchange, as they have historically migrated across different cultures and periods, creating a rich tapestry of storytelling traditions. In particular, we study the aggregated explicit tokens mapped on the values proposed by the Theory of Basic Human Values~\citep{ schwartz1992universals,schwartz2012} across  fairy tale corpora from the traditions of three European countries -- namely Portugal, Italy and Germany -- in order to compare their quantitative representations and analyse the emerging patterns. We do this by first finding the stemmed matches of these tokens and enriching the text with the corresponding annotation. After that we employ a technique called word embedding with a compass~\citep{di_carlo_training_2019} and clique percolations~\citep{palla2005uncovering} to highlight the semantic variation between the three national corpora.

A critical investigation of the results of our method finds that these correspond to findings of previous research. We also find indications that, despite the differences in the expression of values in the three compared countries, it seems that the values of Benevolence (quality of interpersonal relationships), Conformity (respect for social norms and expectations) and Universalism (protection of the welfare of people and nature) have remained consistent in fairy tales across the three national traditions, which we also view as confirmation of the validity of our approach for the study of values embedded in historical, literary corpora.

This paper continues with a background section, introducing key concepts from psychology, literary studies and NLP, and discussing relevant literature from these fields. In Section~\ref{sec:method}, we present our approach and the software tools we used. We provide an overview of our results in Section~\ref{sec:results}. In the final section we conduct a discussion of the approach and results, and we reflect on possible directions for future work.

%% file: sections/020-literature.tex
\section{Background}\label{sec:background}

The study of explicit references of values in fairy tales is related to the accumulated social attitudes up to the historical period of codification of the tales. To our knowledge, no systematic research of this wide topic exists. As such, we view it as being at the crossroads between the socio-historical, literary study of fairy tales, and the psychological study of social values which is shaped by contemporary research. On the other hand, such a study at scale and in a reproducible way would not be possible without the instruments and methods of computational humanities and word embeddings in particular.

\input{sections/021-tales}
\input{sections/022-values}
\input{sections/023-embeddings}

%% file: sections/021-tales.tex
\subsection{Unpacking Fairy-Tale Studies from the Brothers Grimm to Digital Humanities}

The late 18th century witnessed the rise of folklore studies as part of a quest for national and cultural identity, particularly in Europe~\citep{schacker_national_2003}. Jakob and Wilhelm Grimm, riding the tide of renewed interest in popular culture among the upper-class intelligentsia, became pivotal figures in this domain. They first published their fairy-tale collection \textit{Children's and Household Tales} in 1812, striving to present a pure German narrative tradition, untouched by foreign influence, particularly the French~\citep{teverson_fairy_2013}. This publication sparked what would become the 19th century's golden age of fairy tales across Europe. This was a time of growing urbanisation, industrialization, and literacy. Scholars and nationalists, fearful of losing invaluable oral traditions due to these rapid societal changes, began the collection and preservation of folklore~\citep{ostry_social_2013}. Among these custodians were collectors and writers such as Italy's Giuseppe Pitré and Portugal's Consiglieri Pedroso, whose texts feature prominently here alongside the Grimms’. Their work, heavily inspired by the Grimms, was driven by a desire to distil and dialectically construct their nations' cultural legacy. 

Despite the nationalistic intentions of Brothers Grimm and others who embarked on preserving what they thought to be distinct national narratives, the study of fairy tales reveals as much about the interconnectedness of cultures as it does about their uniqueness. Fairy tales, at their core, are a blend of narratives that ``migrate on soft feet''~\citep{warner_once_2016}, indicating that they traverse and interweave across generic, geographical and temporal boundaries, sometimes in untraceable ways. Thus, while the Grimms and others sought to capture and enshrine a uniquely national heritage, their work also serves to underscore the similarities between narrative traditions. 

Unpicking these similarities and differences, however, can prove to be quite a complex task. As scholars are frequently dependent on translations, the risk for misinterpretation or loss of nuanced meanings during this process is high. Translations, like the ones by Margaret Hunt, Thomas Crane and Henriqueta Monteiro used here, are enormously valuable artefacts, but must be recognised as acts of literary adaptation that might differ from the originals~\citep{haase_challenges_2016}. These translations may introduce variations in the representation and interpretation of values, underscoring the need for careful consideration of linguistic nuances in cross-cultural analysis. Further complicating matters, the comparative analysis of several national traditions involves processing vast quantities of text to identify patterns. This challenge extends beyond the study of fairy tales and into the comparative study of literature as a whole.

In response to these challenges, digital humanities and computer-assisted literary studies offer innovative methodologies. Computational methods, in particular, aid in identifying and assessing literary patterns across scales, from individual texts to entire fields and systems of cultural production~\citep{wilkens_digital_2015}. These new approaches, to which our work is a contribution, help produce new types of evidence that enrich and expand humanities research. Indeed, computational approaches to fairy tales have already successfully been deployed in studies such as ``Computational analysis of the body in European fairy tales''~\citep{weingart_computational_2013}. In that study, the authors used digital humanities research methods to analyse the representations of gendered bodies in European fairy tales. They created a manually curated database listing every reference to a body or body part in a selection of 233 fairy tales, and its analysis revealed that the gender and age of fairy-tale protagonists correlate in ways that indicate societal biases, particularly against the ageing female body. A further exploration of gender bias in fairy tales is presented in ``Are Fairy Tales Fair?''~\citep{isaza_are_2023}. This study employs computational analysis to dissect the sequence of events in fairy tales, revealing that one in four event types exhibit gender bias when not considering temporal order, and that female characters are more likely to experience gender-biased events at the start of their narrative arcs. These studies underscore the potential of distant reading, data analysis and visualisation as powerful tools in the comparative study of fairy tales, particularly when used alongside subject expert close reading~\citep{moretti_falso_2022}. Nevertheless, perceptions and attitudes towards gender represent just a fraction of the broader societal values spectrum.

%% file: sections/022-values.tex
\subsection{The expression of values across cultures in European Fairy Tales}

Values are regarded as a shared societal understanding of what constitutes \textit{good}, \textit{wrong}, \textit{fair}, \textit{unfair}, \textit{just}, \textit{right} or \textit{ethical} behaviour~\citep{haidt_righteous_2013,kesebir_morality_2010,turiel_thought_2005}. Values are cognitive representations of an individual's biological needs, an individual’s requirements in interpersonal coordination, and the institutional demands focused on group welfare and survival~\citep{schwartz_toward_1987}. Nonetheless, it is crucial to acknowledge the significance of cultural and individual influence in the development and expression of values. Cultural Psychology postulates that human behaviours result from the reciprocal interaction between cultural and individual psyche~\citep{shweder_thinking_1991,cohen_cultural_2011,schwartz_convergence_2020}. However, the manifestation of behaviours and values is contingent upon context and situation, implying that similar cultural processes might serve or facilitate different purposes based on cultural context~\citep{rogoff_cultural_2003,schwartz_convergence_2020}. Therefore, one could examine variations in the expression of values across different regions and periods, and this could be done through the analysis of historical corpora. This stems from the expectation that literature can be used as a vehicle for the expression of cultural norms and values, thereby reflecting the distinct ideological attributes of the writers and the regions from which it emerges~\citep{albrecht_does_1956}.
Several Theories have been proposed to summarise values across different cultures (for a review of theories see Ellemers et al~\citeyearpar{ellemers_psychology_2019}). In this paper we focus on the Theory of Basic Human Values~\citep{schwartz2012}, since it found validity expression across several cultures~\citep{spini_measurement_2003,schwartz_extending_2001,schwartz_basic_2014}, and it has been applied in the study of European values (e.g., European Social Survey~\citep{davidov_bringing_2008}). 
A version of the Theory of Basic Human Values~\citep{schwartz2012}, simplier than its sequel, comprises of 10 human values that are fuelled by four different and opposite motivations: Openness to Change vs. Conservation, Self-Transcendence vs. Self-Enhancement as observed in Figure~\ref{fig:schwartz}. 

\input{figures-tables/fig010-schwartz}

Openness to Change relates to an individual's need for independence of thought, action, and feelings, and readiness for change, therefore comprises the values of Self-Direction, Stimulation, and partly Hedonism. On the other hand, Conservation relates to the values of Security, Conformity and Tradition, as it emphasises the individual’s needs for order, preservation of the past, and resistance to change. Self-Enhancement considers the individual’s needs to pursue their own interests, success, and dominance over others, therefore comprises the values of Power, Achievement, and partly Hedonism. On the other hand, Self-Transcendence considers the values of Universalism and Benevolence, to focus on  the welfare and better interests of others. For a definition of specific values, see Table~\ref{tab:values}.

\input{figures-tables/tab010-values}

Europeans can be regarded as having a common identity~\citep{castano_2004} that is expressed through their way of life, values and culture, and that has been building since ancient times~\citep{pagden_idea_2002,pinheiro_ideas_2012} leading to the establishment of a broad set of European Values. Values such as human dignity, freedom, democracy, equality, rule of law, and human rights have been declared as the  values of the European Union, to form ``a society in which inclusion, tolerance, justice, solidarity and non-discrimination prevail''~\citep{euvalues}. Based on several empirical studies and policy making guidelines, these values correspond to Schwartz's values of Universalism, Self-Direction, and Benevolence (for more information see~\cite{scharfbillig_values_2021, Murteira_prep}). If these values are presumed to have been shared to some degree across the European territory since antiquity, it stands to reason that they could have been variously conveyed through fairy tales across the three regions under analysis. 

Socio-psychological constructs such as values can either be assessed by explicit or implicit measures. A construct is implicitly assessed when the individual ``is unaware that a psychological measurement is taking place, this type of measure is often used to assess values, attitudes, stereotypes, and emotions in social cognition research''~\citep{apa_dictionary}. On the other hand, a psychological construct is explicitly assessed when the ``individual is aware that a psychological measurement is taking place''~\citep{apa_dictionary}. Putting it simply, values can be measured explicitly when individuals are directly asked about values, and implicitly when the individuals are not aware of the measurement, because values are assessed using indirect questioning methods. Bearing in mind that art is a behavioural expression of culture that serves several purposes, including the \textit{form of order}, which is the need for psychological and mental organisation of experiences~\citep{dissanayake_art_1980}, we can hold the reasonable expectation that the historical corpora under analysis will reflect, to a degree, the explicit and implicit cultural ways and behaviours of societies in which these fairy tales were written.  The presence of these values in our corpora was assessed by quantifying the textual representation employing a word embedding that communicate values in fairy tales. 

One particular type of explicit reference to values, are negative ones, most trivially exemplfied in our corpus of study by ``not loving'' or ``step mother''. However, this notion of opposites to values expands into value dichotomies. These are pairs
of values that are mutually opposed, such as “deceptiveness vs. honesty” or “trust vs.
distrust”. Generally, the alternatives in a duality do not necessarily imply a positive
vs. negative interpretation. To illustrate, none of the options is unequivocally preferable in the dichotomies ``tradition vs. innovation'', ``individualism vs. collectivism'', ``lawfulness vs.
autonomy''~\citep{hardy_values_2022,giouvanopoulou_exploring_2023}. Yet, in the cases when it is a matter of an unambiguously positive value and it's negation, such as ``love vs. hate'' or ``honesty vs. dishonesty'', we argue that the negation of a value is a form of indirect, albeit still explicit, reference to the value itself. Even more, in some cases the two sides of the dichotomy share the same morphological origin. Thus, we argue, that an attempt to capture explicit references to values, also needs to capture negative ones, as is the case when working with vocabulary occurrences.

%% file: figures-tables/fig010-schwartz.tex
\begin{figure}
  \centering

  \includegraphics[width=.6\linewidth]{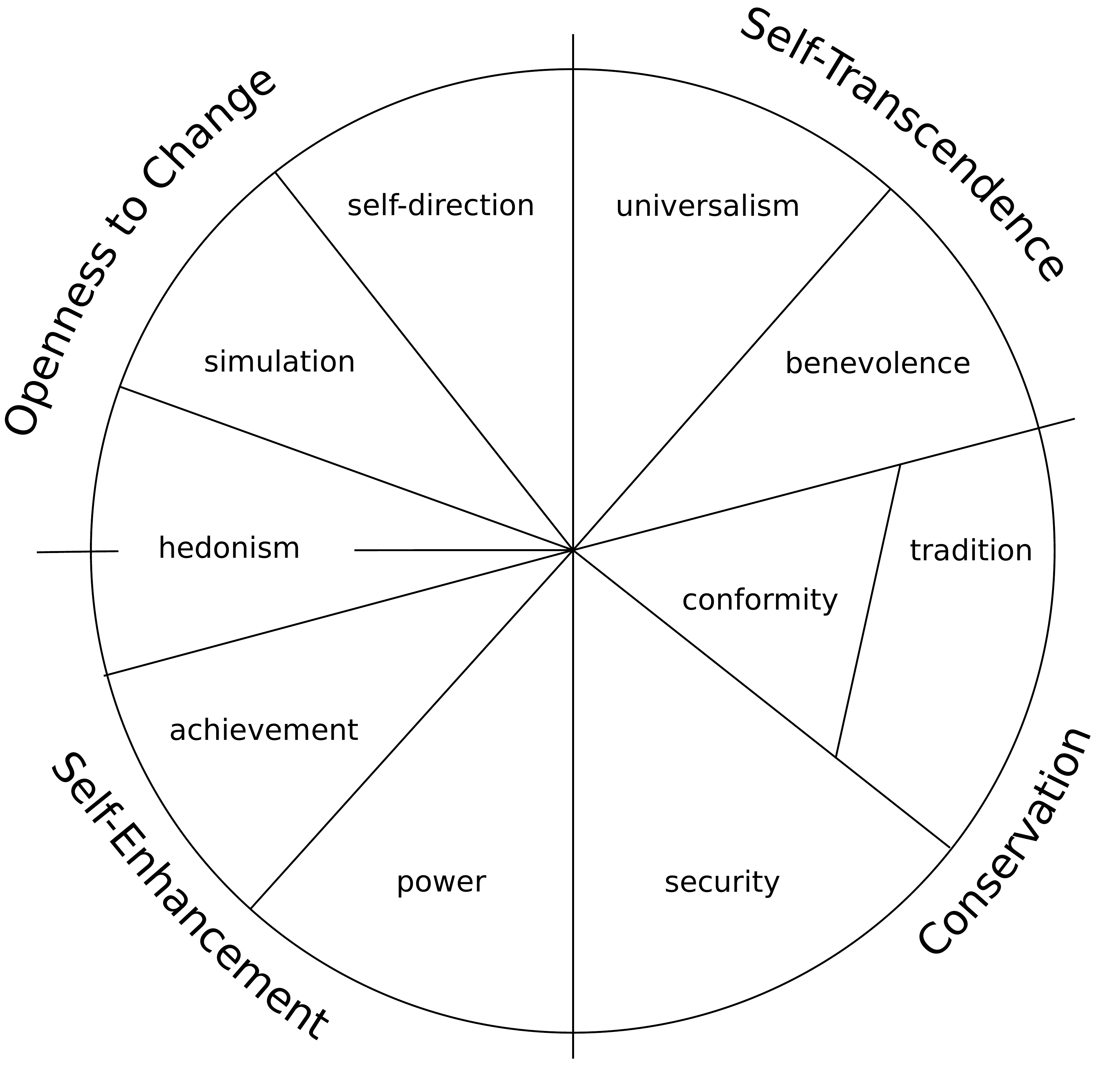}
  
  \caption{Theoretical model of relations among ten motivational types of values~\cite{schwartz2012}.}
  \label{fig:schwartz}
\end{figure}

%% file: figures-tables/tab010-values.tex
\begin{table*}
\centering
\footnotesize
  \begin{tabular}{lp{.8\textwidth}}
    \toprule
Value & Definition \\
    \midrule
Security  & Safety, harmony, and stability of society, of relationships, and of self. \\
Tradition    & Respect, commitment, and acceptance of the customs and ideas that traditional culture or religion provide the self. Maintaining and preserving cultural, family or religious traditions. \\
Conformity & Restraint of actions, inclinations, and impulses likely to upset or harm others and violate social expectations or norms. \\
Self-Direction & Independent thought and action-choosing, creating, exploring. \\
Stimulation & Excitement, novelty, and challenge in life. \\
Hedonism & Pleasure and sensuous gratification for oneself. \\
Achievement & Personal success through demonstrating competence according to social standards. \\
Power & Social status and prestige, control or dominance over people and resources. \\
Benevolence & Preservation and enhancement of the welfare of people with whom one is in frequent personal contact. \\
Universalism & Understanding, appreciation, tolerance, and protection of the welfare of all people and of nature. \\
    \bottomrule
  \end{tabular}
  \caption{The definition of each of the ten motivational types of values~\citep{schwartz2012}.}
  \label{tab:values}
\end{table*}

%% file: sections/023-embeddings.tex
\subsection{Using Word Embeddings to Quantify Vocabulary Differences}

Word embeddings have emerged as an important instrument for the quantitative analysis of textual corpora. These are mappings of vocabulary onto a multidimensional numerical space, based on their occurrences~\citep{mikolov_efficient_2013,rodriguez_word_2022}. Different techniques for creating word embeddings exist, but their common general principle is ``a word is characterised by the company it keeps''. It is useful to distinguish between two categories of word embeddings: i) static (also called type-based) -- those that feature a single numerical representation vector per word token, and ii) contextual (also called token-based) -- those that allow for multiple representations for a word token in order to capture potential nuances in meanings, according to the surrounding context~\citep{miaschi_contextual_2020,lenci_comparative_2022}. Whereas contextual word embeddings better capture the richness of vocabulary, static word embeddings perform better on smaller corpora which do not have the volume that would allow for the semantic richness necessary to represent potential multiple meanings~\citep{ehrmanntraut2021embeddings}. Arguably, this is due to the fact that in a small thematic corpus, typically meanings are restricted by the context of its compilation.

A widespread approach that allows to overcome the challenge of small corpora and their lack of richness, is the combination of pre-training with a huge generic corpus and the subsequent fine-tuning with the corpus of interest. For example, the most popular contextual language model BERT is trained on a corpus that includes the entire contents of Wikipedia which comprises of ~2.5 billion word tokens~\citep{devlin_bert_2019}, others use training sets that are many orders of magnitude larger~\citep{dodge_documenting_2021} However, corpora of these huge dimensions are inevitably contemporarily written, and due to cultural and linguistic change over time inevitably introduce unwanted biases~\citep{de-vassimon-manela-etal-2021-stereotype,ahn-oh-2021-mitigating,mozafari_hate_2020,cuscito_how_2024}. In confirmation of this consideration, particularly forp the context of Historical English, Manjavacas and Fonteyn~\citeyearpar{manjavacas_adapting_2022} observed that training from the ground up is more effective than fine-tuning of preexisting models, and this has been independently confirmed by Cuscito et al~\citeyearpar{cuscito_how_2024}.

When it comes to comparing the word embeddings representing different corpora, a widespread approach is the so-called semantic change detection~\citep{tahmasebi_2021_5040241}. Since for intercultural comparison, ``change'' might wrongly suggest a (diachronic) transition from one culture to the other, when comparing contexts that are not sequential, a more appropriate wording in this context is (synchronic) ``semantic variation''~\citep{tahmasebi_2021_5040241,schlechtweg_wind_2019}. Still, whenever techniques for semantic change detection do not rely on any particular diachronic properties of the underlying corpora, we claim they could be reused also for synchronic linguistic analysis. More specifically we claim that an approach called temporal word embedding with a compass~\citep{di_carlo_training_2019} is applicable, for culture-specific rather than time-specific distinctive corpora. This approach consists of first creating an embedding on a cumulative corpus containing all texts from the different cultures to be considered. Then, from this baseline (compass) word embedding, further fine-tuning is performed on each of the corpora, to be compared so as to create culture-specific word embeddings. The result for each corpus is a different vector representations of each particular word token, which allows for quantitative comparisons between them, as done previously~\citep{ferrara2022ai4ch,di_carlo_training_2019}.

%% file: sections/030-method.tex
\section{Method}\label{sec:method}

To describe our method, we focus first on the followed process, and then on the bespoke tool that was developed to facilitate this process.

\subsection{Process}\label{sec:process}

\input{figures-tables/fig020-process}

Our study of the explicit references to values in fairy tales follows the process illustrated in Figure~\ref{fig:process}. To provide an outline, it starts with the identification of tokens that represent values of interest. We group these tokens in groups that we consider to be synonyms in the studied context. Then, we automatically annotate all occurrences in the text of the stems representing the considered tokens. Once this is done, we manually analyse the produced annotations to identify ambiguities and mistakes in this token identification process. The purpose of this analysis is to better understand the semantics behind their occurrences, in order to refine the selection of tokens and identify potential ambiguities arising from a single syntactical token potentially representing multiple values. Finally, we apply a static word embedding with a compass and perform critical analysis on the differences and similarities from the resulting vector spaces.

\input{figures-tables/tab020-corpora}

\paragraph{Fairy Tales Corpora} The corpus selection had several stages. First, we focused on the Grimms’ \textit{Children’s and Household Tales}, using Margaret Hunt’s 1884 English translation. We manually selected 30 tales that span well-known and beloved stories and lesser-known ones, so as to provide a comprehensive representation of the entire collection. Then we selected 30 Portuguese and 30 Italian tales taken from two important contemporary collections to the Grimms’: \textit{Portuguese Folktales} by Consiglieri Pedroso, translated to English in 1882 by Henriqueta Monteiro; and \textit{Italian Popular Tales}, collected and translated to English in 1885 by Thomas Frederick Crane. These collections were chosen due to their cultural significance and their temporal proximity to the Grimms' collection, aiming to offer a comparative perspective on 19th century fairy tales across different European cultures.

\paragraph{Selection of Tokens} Assuming that the historical corpora are themselves mirrors of social behaviours and ways of living in societies in which the fairy tales were written, we are interested in the explicit expressions of values in the texts. Starting from Schwartz's model and the European core values, we initially compose a list of tokens that represent these values, based on three empirical studies regarding value-specific tokens. This list of tokens contains words that were selected from two dictionary studies about values, where each word is associated with a specific value from the 10 identified by Schwartz.~\citep{schwartz1992universals,lindeman_measuring_2005, Murteira_prep}. For instance, the token “Peace” is associated with the value of Universalism, and the token “Cooperation” is associated with the value of Benevolence
(see Table~\ref{tab:tokens} in Appendix).
Then we perform automatic identification of explicit references of these tokens and relate them to the corresponding values. We do this using stemming~\citep{jabbar_empirical_2020} on both the token lists and the fairy tale texts. This is because, in contrast e.g. to lemmatisation, stemming reduces different word forms to the same originating token. We use the Snowball stemmer algorithm~\citep{porter2001} to identify all occurrences of the stemmed tokens in the corpora and tag (i.e. annotate) them with a label corresponding to the group of synonym tokens. 

\paragraph{Critical Review} We then critically analyse and refine by adapting tokens according to the desired annotation. This was done using a graphical interface that was specifically developed for the purpose and allows for a review of the texts in the corpora with the results of the automatic annotation highlighted in different colours. The tool is discussed in more details in Section~\ref{sec:tools}. The outcome of this was a series of decisions to adjust the token selection as a way to refine it and guide subsequent iterations of this annotation process. Correspondingly, following this approach inspired by grounded theory~\citep{rieger_discriminating_2019}, the ultimately proposed list of tokens in this study emerges from exploration of the corpus and is not a result of deductive hypothesis research. We provide a statistical overview of the results of this annotation process in Table~\ref{tab:corpora} and a Venn diagram of the occurrences of groups of synonym token across the three corpora in Figure~\ref{fig:venn-tokens}. Furthermore, in Appendix we provide the complete final version of our tokens.

\input{figures-tables/fig097-venn-tokens}

\paragraph{Word Embedding with a Compass} Due to the historical nature of the studied corpora and in order to avoid contaminating them with external biases from pre-training, we organise our analysis following the word embedding with a compass approach~\citep{di_carlo_training_2019}. To do this, we create one generic culture-agnostic shared embedding from scratch containing all three corpora. Then, starting from this compass, we independently create three parallel fine-tunings for each of the cultures. For the creation of the compass, to avoid the possible introduction of biases, we chose not to include any further possible texts, neither from any of our three contexts, nor from others. Our approach to syntactic identification of references of values, is not contextual, i.e. we treat a reference to a value-related stemmed token as the same for all its identified uses. This is why, in our critical review step, we examined the validity of this generalisation. To represent the annotations in the word embedding algorithm, before and after each identified occurrence of a token we insert an indication of the corresponding group of synonym tokens (i.e. the first token in that group).

\paragraph{Comparison of Semantic Variation} The word embedding allows measuring contextual similarity between words, thus speaking of ``change'' and ``variation''. Once we have the three word embeddings for the cultural corpora, for each of them we consider only the distances between groups of tokens (represented by the annotation label, i.e. the first token in each synonym group) and experimentally define a similarity threshold above which we consider a pair of tokens to have a relating edge between them in a graph representation of tokens) in order to use clique percolations clustering with k=2~\citep{palla2005uncovering}. In other words, for all similarities above that threshold we consider the corresponding tokens to be related in that embedding, and distances above the threshold mean the corresponding tokens are not. This results in a clustering that might assign one token to multiple clusters. It might also bring two tokens into the same cluster even if the distance between them is greater than the threshold, as long as there is a ``bridge'' of other tokens in between to connect them.

\paragraph{Historical and Social Critical Analysis} At the end of our method, we analyse the quantitative results using critical analysis from the perspectives of both literary studies and psychological research. This allows us to cross-validate (e.g. through  triangulation~\citep{noble_triangulation_2019}) our results with the established body of research and thus get an indication of their theoretical validity.

\subsection{Automated Annotation Tool}\label{sec:tools}

To facilitate the critical analysis of the annotations, we developed a bespoke tool -- named \textsc{MoreEver}\footnote{Accessible online at \site} -- that automatically identifies the explicit references to values, highlights them for critical human review of the tales and annotations, and provides some simple visualisation techniques to ease the comparative analysis. The main view of the annotation tool is provided in Figure~\ref{fig:screenshot}.

\input{figures-tables/fig030-screenshot}

Both texts titles on the left and tokens on the right are clickable, which allows easy browsing per corpus to explore individual fairy tales, as well as per value token.
Through a dropdown box visible in its upper left corner of Figure~\ref{fig:screenshot}, the tool features a list of vocabulary generalisation techniques, intended as techniques to identify a broader range of tokens as matching. The choice of the technique in use can be changed in real time to allow users to examine in context throughout the corpora texts which vocabulary generalisation best approximates the expression of values they are aiming for. Among these generalisation techniques are lemmatisation, as well as Porter~\citep{porter_algorithm_2006}, Snowball~\citep{porter2001} and Lancaster~\citep{paice_another_1990} stemmers\footnote{See implementations by \url{https://www.nltk.org/api/nltk.stem.html}}. The tool further supports no reduction (i.e. identification only of the exact matching words) and repeated application of Snowball stemmer for experimental purposes. This feature allowed our research team to make an informed choice for the use of Porter's Snowball stemmer.

On top of these features, \textsc{MoreEver} provides basic functionalities for interactive exploratory visualisations in the form of heatmap and Venn diagrams. Heatmaps, as shown in Figure~\ref{fig:heatmap_detailed} provide a bird's-eye view of the occurrences of tokens in tales. The featured 3-set Venn diagrams provide a cross-section of the occurrences of tokens across the three national corpora, as seen in Figure~\ref{fig:venn-tokens}. Both visualisations are dependent on the choice of vocabulary generalisation technique, and the provided examples are derived for the Snowball stemmer as the technique of choice in this paper.

\input{figures-tables/fig098-heatmap-texts-labels}

%% file: figures-tables/fig020-process.tex
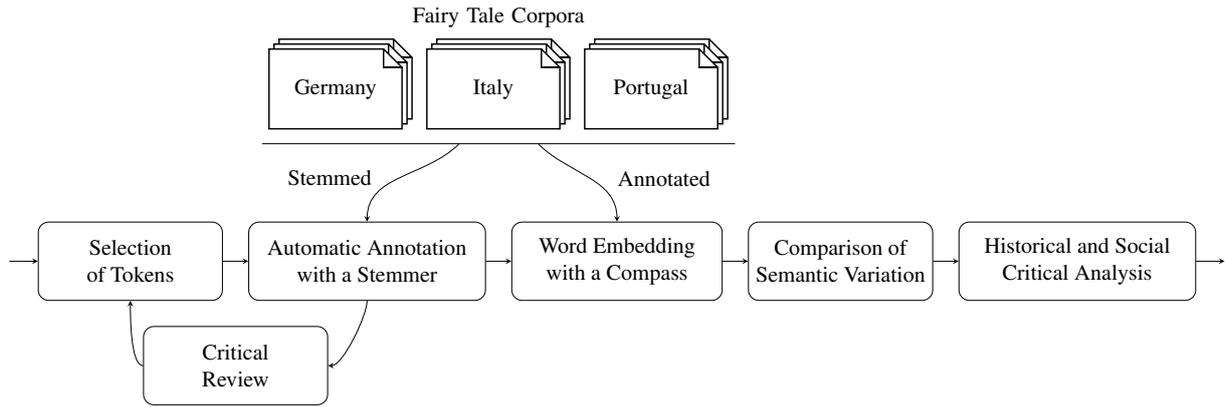
\begin{figure*}
  \centering
\begin{adjustbox}{width=\textwidth, keepaspectratio}  
\begin{tikzpicture}
\begin{scope}[shift={(6.4,3)}]
    \node[text width=7cm, align=center, anchor=north] at (0.1, 2) {Fairy Tale Corpora};
 
    \node[rect={}{}{}{solid},
        align=center,
        minimum width=9cm,
        minimum height=1.5cm,
        ] (Corpora) at (0.1,0) {};

    \node[doc,
        minimum width=2.5cm,
        minimum height=1.5cm,
        fill=white,
        double copy shadow] (DocsA) at (-3,0.3) {Germany};

    \node[doc,
        minimum width=2.5cm,
        minimum height=1.5cm,
        fill=white,
        double copy shadow] (DocsB) at (0,0.3) {Italy};

    \node[doc,
        minimum width=2.5cm,
        minimum height=1.5cm,
        fill=white,
        double copy shadow] (DocsC) at (3,0.3) {Portugal};
\end{scope}

\begin{scope}[shift={(-6,0)}] (Values)


    \node[draw,
        align=center,
        minimum width=3.5cm, 
        minimum height=1.5cm,
		rounded corners=0.2cm] (Labels) at (5.5,0) {Selection\\of Tokens};

\end{scope}

    \node[draw,
        align=center,
        minimum width=4.5cm, 
        minimum height=1.5cm,
		rounded corners=0.2cm] (Annotation) at (4,0) {Automatic Annotation\\with a Stemmer};
 
    \node[draw,
        align=center,
        minimum width=4cm, 
        minimum height=1.5cm,
		rounded corners=0.2cm] (Embedding) at (8.75,0) {Word Embedding\\with a Compass};
 
    \node[draw,
        align=center,
        minimum width=3.5cm, 
        minimum height=1.5cm,
		rounded corners=0.2cm] (Tune) at (13,0) {Comparison of\\Semantic Variation};
 
    \node[draw,
        align=center,
        minimum width=4.5cm, 
        minimum height=1.5cm,
		rounded corners=0.2cm] (Ana) at (17.5,0) {Historical and Social\\Critical Analysis};
 
    \node[draw,
        align=center,
        minimum width=3.5cm, 
        minimum height=1.5cm,
		rounded corners=0.2cm] (Review) at (1.5,-2) {Critical\\Review};
 

 \draw[-stealth] (-2.8,0) -- (Labels);

\draw[-stealth] (Labels) -- (Annotation);

\draw[-stealth] (Annotation) edge[
	out=south,
	in=east,
	looseness=0.5] node[anchor=north]{} (Review);

\draw[-stealth] (Review) edge[
	out=west,
	in=south,
	looseness=0.5] node[anchor=north]{} (Labels);

\draw[-stealth] (Annotation) -- (Embedding);

\draw[-stealth] (Corpora) edge[out=-135, in=90] node[anchor=east,shift={(-0.4,0)}]{Stemmed} (Annotation);

\draw[-stealth] (Corpora) edge[out=-45, in=90] node[anchor=west,shift={(0.4,0)}]{Annotated} (Embedding);

\draw[-stealth] (Embedding) edge node[anchor=south]{} (Tune) ;

\draw[-stealth] (Tune)  edge node[anchor=south]{} (Ana);

\draw[-stealth] (Ana) -- (20.3,0);

\end{tikzpicture}
\end{adjustbox}
  \caption{The outline of the process we followed.}
  \label{fig:process}
\end{figure*}

%% file: figures-tables/tab020-corpora.tex
\begin{table}[b]
\centering
\footnotesize
  \begin{tabular}{lrrrr}
    \toprule
Corpus & Texts & Symbols & Words & Tokens \\
    \midrule
Germany (1884)  & 30 & 306 475 & 59 500 & 1840 \\
Italy (1885)    & 30 & 234 158 & 45 223 & 1808 \\
Portugal (1882) & 30 & 231 149 & 44 887 & 1439 \\
    \bottomrule
  \end{tabular}
\caption{Descriptive statistics of the corpora. When we refer to tokens, we mean the ones that were identified by our automated annotation process.
Complete list of included texts is available in Table~\ref{tab:texts} in Appendix.}
  \label{tab:corpora}
\end{table}

%% file: figures-tables/fig097-venn-tokens.tex
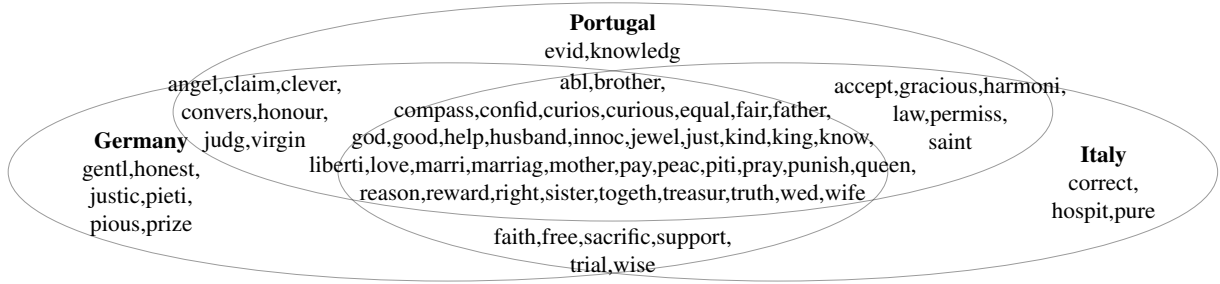
\begin{figure*}[!ht]
\centering
\resizebox{\textwidth}{!}{

\begin{tikzpicture} [set/.style = {
    circle,
    minimum size = 9cm,
    opacity = 0.4,
    text opacity = 1}]

\draw[set] (-3,0) ellipse (8cm and 2cm);
\draw[set] (3,0) ellipse (8cm and 2cm);
\draw[set] (0,1.1) ellipse (8cm and 2cm);

\node (A) [anchor=east, align=center] at (-4,0.5) {};
\node (C) [anchor=west, align=center] at (4,0) {};
\node (B) [anchor=south, align=center] at (0,4) {};

\node at (barycentric cs:A=1,B=-.15,C=-.3) [left, align=center, anchor=west] {\textbf{Germany}\\gentl,honest,\\justic,pieti,\\pious,prize};
\node at (barycentric cs:A=1,B=5,C=1) [below, align=center, anchor=north] {\textbf{Portugal}\\evid,knowledg};
\node at (barycentric cs:A=-.47,B=0.02,C=1.1) [right, align=center, anchor=east] {\textbf{Italy}\\correct,\\hospit,pure};
 
\node at (barycentric cs:A=-95,B=7,C=3) [left, align=center, anchor=south east] {angel,claim,clever,\\convers,honour,\\judg,virgin};
\node at (barycentric cs:A=1,B=-.45,C=1) [below, align=center] {faith,free,sacrific,support,\\trial,wise};
\node at (barycentric cs:A=0,B=.7,C=10) [right, align=center, anchor=south west] {accept,gracious,harmoni,\\law,permiss,\\saint};
\node at (barycentric cs:A=1,B=.2,C=1) [align=center] {abl,brother,\\compass,confid,curios,curious,equal,fair,father,\\god,good,help,husband,innoc,jewel,just,kind,king,know,\\liberti,love,marri,marriag,mother,pay,peac,piti,pray,punish,queen,\\reason,reward,right,sister,togeth,treasur,truth,wed,wife}; %
 
\end{tikzpicture}}
  \caption{A Venn diagram showing the occurrences of stemmed tokens across the national corpora.}
  \label{fig:venn-tokens}
\end{figure*}

%% file: figures-tables/fig030-screenshot.tex
\begin{figure*}
  \centering
  \includegraphics[width=\textwidth]{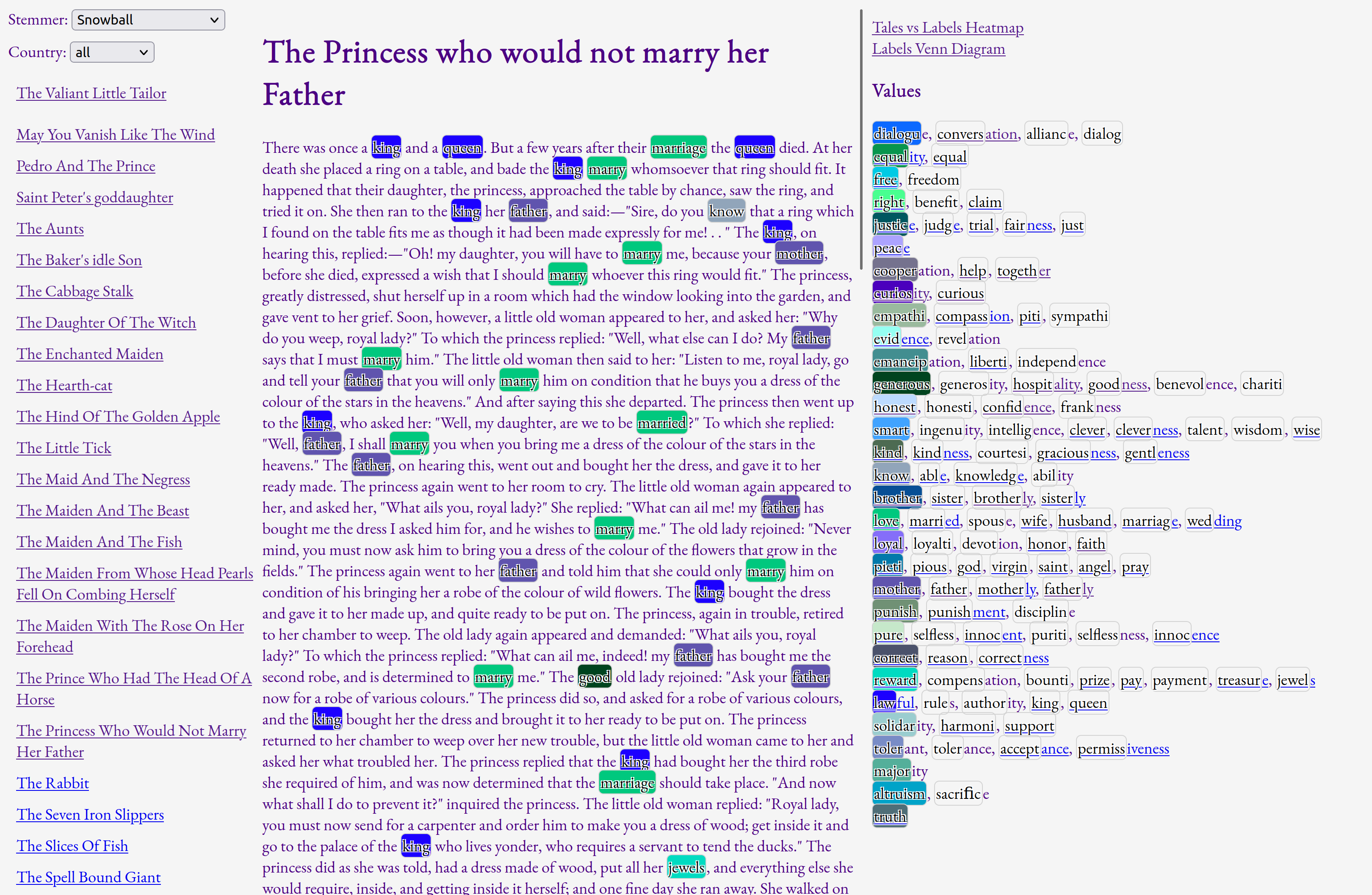}
  \caption{Screenshot of a browsing page from \textsc{MoreEver}, the bespoke web instrument that reviews the produced annotations.
  Another view shows a clickable heatmap as Figure~\ref{fig:heatmap_detailed} in Appendix, which allows for a distant reading view.
  }
  \label{fig:screenshot}
\end{figure*}

%% file: figures-tables/fig098-heatmap-texts-labels.tex
\begin{figure*}
  \centering
  \includegraphics[width=\textwidth]{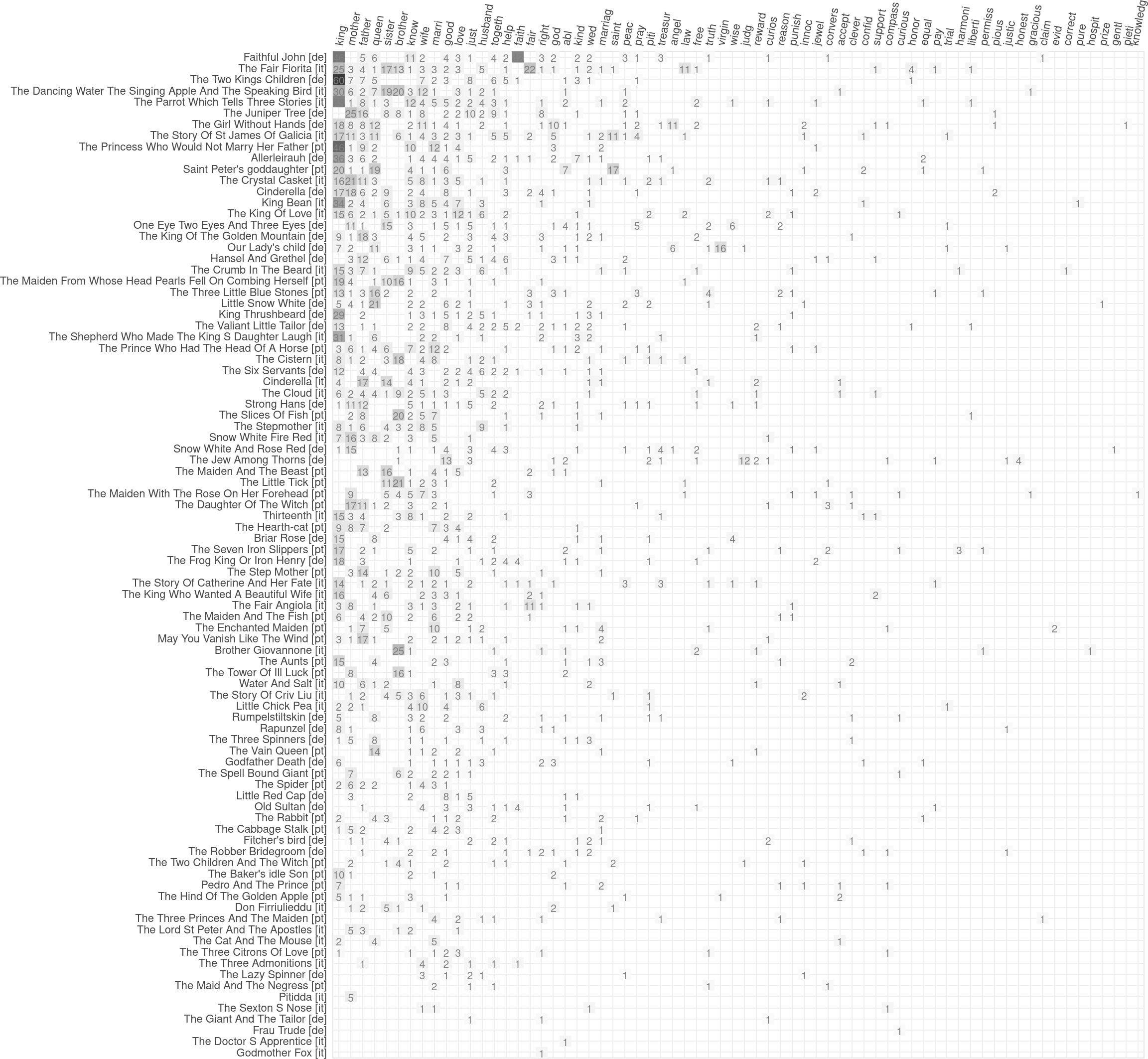}
  \caption{Counts of identified occurrences of stemmed tokens across the texts of the three corpora. An interactive version of this heatmap is available in \textsc{MoreEver}. In it clicking on a number takes you to the corresponding text for easier review.}
  \label{fig:heatmap_detailed}
\end{figure*}

%% file: sections/040-results.tex
\section{Results}\label{sec:results}

\input{figures-tables/fig040-heatmap-countries-values}

An important part of the results of our approach is the reflective inspection of the produced automated annotation and possible corrections for these. An overall conclusion of this process is that, expectedly, the most impactful tokens capture the values  they were intended to match well. The most important token that did not correspond to our initial interpretation was ``faith''. We originally ascribed the label ``faith'' to the value of ``piety'', indicating religious devotion. However, a careful examination of our corpus revealed an intriguing trend. The term ``faith'', contrary to our initial classification, exclusively expressed affiliation with ``loyalty'', mainly as per the usage patterns in various Grimm tales, particularly in ``Faithful Johannes'' (a German tale). As a consequence, we ascribe the token ``faith'' to the value associated to ``loyalty''.

Another token that provides an interesting example is ``father'', due to its potential multiple associations. On the one hand, it could represent ``caring'',  similar to ``mother'', but on the other, it could be a symbol of authority~\citep{hopp2021extended}. When exploring the corpora, we found that ``father'' was predominantly associated with ``caring'', with a remarkable exception in ``The Maiden and the Fish'' (Portugal), where one out of four instances appeared associated with authoritative power.

A third, less impactful token we considered was ``patient'', which was initially intended as associated with ``patience'' and ``kindness''. However, an analysis of the corpus found that its usage related exclusively to an individual receiving medical treatment, and we consequently excluded it from our analysis.

Figure~\ref{fig:heatmap_aggregated} shows the references to values by countries, according to the ascribed tokens.
A more detailed mapping of occurrences of tokens in particular texts is provided in  Figure~\ref{fig:heatmap_detailed} in the Appendix.
From the resulting comparison of clusters across corpora, noteworthy is the one defined around tokens related to ``mother''. As the Venn diagram on Figure~\ref{fig:venn-clusters} shows, while in our German and Portuguese corpora this token of reference appears together with ``brother", in the Italian and Portuguese corpora, it also appears in relation to ``know''. Only in Germany does it relate to ``generous". Noteworthy, despite our previous comment regarding ``father'', this token does not appear in the cluster.

\input{figures-tables/fig050-venn-clusters}

\subsection{Historical Analysis}

Dolores Buttry elucidates on the usage of ``faith'' in Grimm tales to exclusively mean ``loyalty'', and not ``piety''. She writes that the related values of faithfulness and loyalty (which are ``Treu'' and ``Treue'' in German) have been foundational virtues in Germany since ancient times~\citep{buttry_treue_2011}. Stories such as ``Faithful Johannes'', but also ``The Frog King'', exemplify extreme loyalty towards superiors, illustrating the importance of fidelity and respect for authority in their various manifestations. Buttry characterises the tale of the loyal servant as an enduring archetype, highlighting the recurring appearance of the words ``Treu'' (faithful) and ``Treue'' (loyalty, fidelity) in German tales~\citep{buttry_treue_2011}. She further suggests that, while respect for authority and the sanctity of oaths were nearly universal concepts before these stories were collected, they seem to have retained their vitality and cultural significance particularly in German-speaking traditions. This idea finds further support in one of the only non-German occurrences of ``faith'' in our corpus, as the label appears in ``The Story Of Catherine and Her Fate,'' a Sicilian tale first collected by Swiss-German folklorist Laura Gozenbach.

It is also interesting to examine how values manifest in tales from different cultural contexts. In our results, we found that values of ``piety'' and ``empathy'' appeared clustered together in Italian and Portuguese tales, but not in German ones. This may be explained by the different religious traditions in all three countries, since both Italy and Portugal were predominantly Catholic regions at the time the tales were collected, while there was a strong Protestant presence in the German territory. Indeed, Jack Zipes~\citeyearpar{zipes_brothers_2002} writes that the Grimms' tales portrayed the main values of Protestant ethics and the bourgeois enlightenment. The heroes in their tales are predominantly concerned with self-preservation and the acquisition of wealth, and  they assist others, including animals, only when they perceive a potential gain for themselves, demonstrating a calculated approach to empathy and compassion. This model of behaviour, Zipes argues, exemplifies the general Protestant ethic of the time, and so empathy, although occasionally appearing in the Grimms' tales, is not a dominant theme~\citep{zipes_brothers_2002}. We may advance the possibility that the differing religious ethos of Italy and Portugal would place more emphasis on empathy as it relates to Catholic piety.

\subsection{Social Analysis}

Frequency analysis shows that tokens such as ``mother'', ``law'', brother'', and ``love'' have a strong presence (more than 100 appearances, see Figure~\ref{fig:heatmap_aggregated}) across the three countries under analysis. Based on the elaborated correspondence between tokens and the Theory of Basic Values (see Appendix), the words ``mother'', ``brother'' and ``love'' are connected to Benevolence, and ``law'' is connected to Conformity. In Germany, the token ``justice'' has also a strong presence and is connected with the value of Universalism which stands for the protection and welfare of all people and nature. Considering that the value Benevolence stands for the good quality of social connections between people, and Conformity stands for the preservation of socio-cultural expectations and norms, then we could infer that these tales describe several social dynamics.  The tales’ plots are representative of dynamics among fictional characters that may resemble society, in order to describe the quality of human relationships and socio-cultural norms in place.

Interestingly, some differences across countries are expressed by the token frequencies related to Benevolence, Conformity and Universalism. For instance, in Germany, ``mother'' seems to be a stronger reference for communication of Benevolence than ``brother'' when compared to Portugal and Italy. Also, ``love'' seems to be a stronger reference for communication of Benevolence in Italy than in Germany and Portugal. However, in Germany, we may note that tokens such as ``generous'' and ``cooperation'' reinforce the communication and expression of Benevolence in those tales. Turning to the need for rules and social welfare, it seems that in Germany and Italy the token ``law'' is frequently used when compared to Portugal to express the value of Conformity. Finally, the German corpus shows a strong presence of the token ``justice'' in their tales, which highlights the importance of Universalism in this context and the need to convey the respect for human rights and dignity. In sum, while Portugal, Italy and Germany communicate strongly the values of Benevolence and Conformity, it seems that Germany also communicates the value of Universalism. Despite these nuances, it seems that European Values of Benevolence and Universalism are being communicated by the tales across all three countries.

%% file: figures-tables/fig040-heatmap-countries-values.tex
\begin{figure*}
  \centering
  \includegraphics[width=\textwidth]{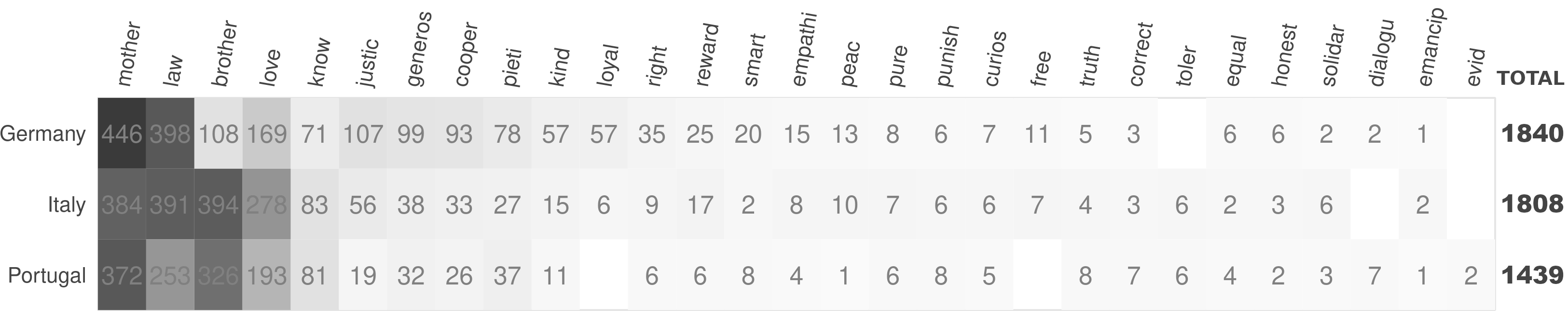}
  \caption{Frequencies of identified occurrences of stemmed tokens across the three corpora.
  A more detailed heatmap between texts and labels is available on Figure~\ref{fig:heatmap_detailed} in Appendix.
  }
  \label{fig:heatmap_aggregated}
\end{figure*}

%% file: figures-tables/fig050-venn-clusters.tex
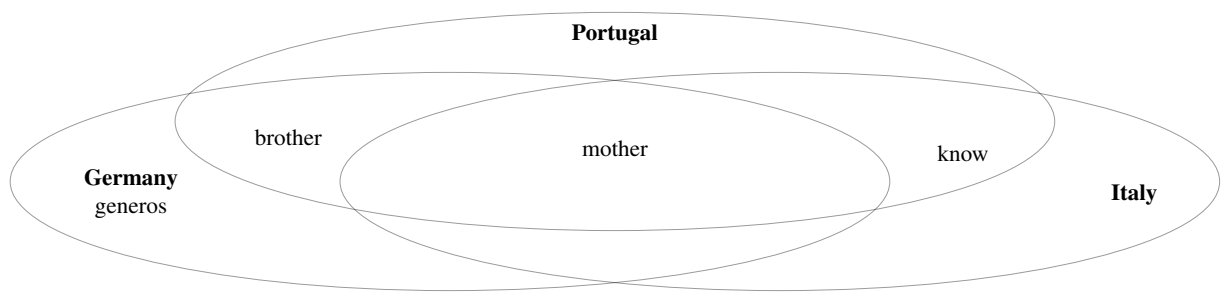
\begin{figure*}
\centering
\resizebox{\textwidth}{!}{

\begin{tikzpicture} [set/.style = {
    circle,
    minimum size = 9cm,
    opacity = 0.4,
    text opacity = 1}]

\draw[set] (-3,0) ellipse (8cm and 2cm);
\draw[set] (3,0) ellipse (8cm and 2cm);
\draw[set] (0,1.1) ellipse (8cm and 2cm);

\node (A) [anchor=east, align=center] at (-4,0.5) {};
\node (C) [anchor=west, align=center] at (4,0) {};
\node (B) [anchor=south, align=center] at (0,4) {};

\node at (barycentric cs:A=1,B=-.15,C=-.3) [left, align=center, anchor=west] {\textbf{Germany}\\generos};
\node at (barycentric cs:A=1,B=5,C=1) [below, align=center, anchor=north] {\textbf{Portugal}};
\node at (barycentric cs:A=-.47,B=0.02,C=1.1) [right, align=center, anchor=east] {\textbf{Italy}};
 
\node at (barycentric cs:A=-95,B=1,C=10) [left, align=center, anchor=south east] {brother};
\node at (barycentric cs:A=1,B=-.45,C=1) [below, align=center] {};
\node at (barycentric cs:A=-2,B=.7,C=10) [right, align=center, anchor=south west] {know};
\node at (barycentric cs:A=1,B=.2,C=1) [align=center] {mother}; %
 
\end{tikzpicture}}
  \caption{An illustration of the degree of overlap across the three national corpora for the token ``mother''. Note that the visible tokens are stemmed.}
  \label{fig:venn-clusters}
\end{figure*}

%% file: sections/050-conclusion.tex
\section{Discussion and Conclusion}\label{sec:conclusion}

While the provided analysis is open-ended, and the emerging results would require more thorough examination, our early analysis provides some concrete evidence that European Values have been a long-standing element in European cultural communication through fairy tales. The corpus analysis across different cultures revealed a significant variety in the representation of values. For example, the affiliation of the token "faith" with "loyalty" rather than "piety," particularly in German culture, illustrates the role of cultural and historical contexts in shaping value representations. Similarly, the differential clustering of "piety" and "empathy" in Italian and Portuguese tales compared to German tales further underscores the influence of religious and socio-cultural contexts in value representation. Interestingly, despite these differences, the analysis revealed a strong commonality across all three cultures, pointing at the communication of European Values through tales. Tokens associated with Benevolence, Conformity, and Universalism manifested frequently across fairy tales of all three countries. This finding is particularly noteworthy because it suggests a strong shared cultural understanding and expression of these values across European literary production, and, possibly and by extension, across European societies, thus hinting at the existence of a pan-European cultural memory.

We have identified clear limitations in our approach. Working at the syntactic level, both in terms of stemming and static word embeddings, limits the possibility to capture nuances, and with this some noise is introduced in the analysis. However, contrary to our expectations, our detailed analysis by means of in-depth close reading revealed that ambiguities are rather a noteworthy exception and not the norm. This is valid to the extent that in none of these cases a token bore semantic ambiguity that was a dichotomy rather than an outlier so that it could undermine the general results.

The focus on explicit references, unsurprisingly, resulted in an inability to annotate tokens such as “democracy” in the tales, as they were only implicitly referenced. Contextual language models are also able to capture indirect relatedness of from the context~\citep{montanelli_survey_2023}. This has also been attempted in the context of values, notably in the ValueEval competition~\citep{kiesel_semeval-2023_2023,ferrara_augustine_2023,papadopoulos-etal-2023-andronicus}. However, such approaches are undermined by the variance of value perceptions among humans. Efforts to annotate values for a ground truth chronically suffer from appalling agreement rates. In particular, when employing an even number of annotators, Hoover and colleagues report ties (disagreement) above 60\% across more than 6000 tweets~\citep{hoover_moral_2020}. When annotating arguments for values, Kiesel and colleagues report agreement with Krippendorff's $\alpha$ of 0.49~\citep{kiesel_identifying_2022}, which is well below the 0.667 that Krippendorff calls ``the lowest conceivable limit''~\citep{krippendorff_reliability_2004}. 
Furthermore, when using contextual word embeddings, due to the needed corpus sizes, an approach that combines of pre-training and fine-tuning becomes necessary.
Considering this, we believe special attention should be paid to the possibility that the pre-trained embeddings may introduce biases unrelated to the corpus under study.

This work provides a foundational understanding of how European Values are represented in literary texts and highlights the potential of computational linguistics in cultural studies. This study encourages further interdisciplinary research in the field of literary studies, cultural analytics, and computational linguistics to expand our understanding of cultural values and their historical evolution.

%% file: figures-tables/tab090-texts.tex
\begin{table*}[!htbp]
\footnotesize
  \begin{tabular}{p{.25\textwidth}p{.35\textwidth}p{.33\textwidth}}
    \toprule
    \makecell{Germany\\~\cite{grimm2022household}} & \makecell{Italy\\~\cite{crane2017popular}} & \makecell{Portugal\\~\cite{consiglieri2017folktales}} \\
    \midrule
\begin{tabitemize}
\item Allerleirauh
\item Briar Rose
\item Cinderella
\item Faithful John
\item Fitcher's Bird
\item Frau Trude
\item Godfather Death
\item Hansel And Grethel
\item King Thrushbeard
\item Little Red Cap
\item Little Snow White
\item Old Sultan
\item One Eye Two Eyes And Three Eyes
\item Our Lady's Child
\item Rapunzel
\item Rumpelstiltskin
\item Snow White And Rose Red
\item Strong Hans
\item The Frog King Or Iron Henry
\item The Giant And The Tailor
\item The Girl Without Hands
\item The Jew Among Thorns
\item The Juniper Tree
\item The King Of The Golden Mountain
\item The Lazy Spinner
\item The Robber Bridegroom
\item The Six Servants
\item The Three Spinners
\item The Two Kings Children
\item The Valiant Little Tailor    
\end{tabitemize}

& 

\begin{tabitemize}
\item Brother Giovannone
\item Cinderella\textsuperscript{\ref{editor:imbriani}}
\item Don Firriulieddu
\item Godmother Fox
\item King Bean Giuseppe Bernoni
\item Little Chick Pea Tuscan variant
\item Pitidda
\item Snow White Fire Red
\item The Cat And The Mouse
\item The Cistern
\item The Cloud\textsuperscript{\ref{editor:comparetti}}
\item The Crumb In The Beard\textsuperscript{\ref{editor:coronedi}}
\item The Crystal Casket
\item The Dancing Water The Singing Apple And The Speaking Bird
\item The Doctor's Apprentice
\item The Fair Angiola\textsuperscript{\ref{editor:gozenbach}}
\item The Fair Fiorita\textsuperscript{\ref{editor:comparetti}}
\item The King Of Love
\item The King Who Wanted A Beautiful Wife\textsuperscript{\ref{editor:gozenbach}}
\item The Lord St Peter And The Apostles
\item The Parrot Which Tells Three Stories
\item The Sexton's Nose
\item The Shepherd Who Made The King's Daughter Laugh\textsuperscript{\ref{editor:gozenbach}}
\item The Stepmother
\item The Story Of Catherine And Her Fate\textsuperscript{\ref{editor:gozenbach}}
\item The Story Of Crivoliu\textsuperscript{\ref{editor:gozenbach}}
\item The Story Of St James Of Galicia\textsuperscript{\ref{editor:gozenbach}}
\item The Three Admonitions
\item Thirteenth
\item Water And Salt
\end{tabitemize}
&

\begin{tabitemize}
\item May You Vanish Like The Wind
\item Pedro And The Prince
\item Saint Peter's Goddaughter
\item The Aunts
\item The Baker's Idle Son
\item The Cabbage Stalk
\item The Daughter Of The Witch
\item The Enchanted Maiden
\item The Hearth-cat
\item The Hind Of The Golden Apple
\item The Little Tick
\item The Maid And The Negress
\item The Maiden And The Beast
\item The Maiden And The Fish
\item The Maiden From Whose Head Pearls Fell On Combing Herself
\item The Maiden With The Rose On Her Forehead
\item The Prince Who Had The Head Of A Horse
\item The Princess Who Would Not Marry Her Father
\item The Rabbit
\item The Seven Iron Slippers
\item The Slices Of Fish
\item The Spell Bound Giant
\item The Spider
\item The Step Mother
\item The Three Citrons Of Love
\item The Three Little Blue Stones
\item The Three Princes And The Maiden
\item The Tower Of Ill Luck
\item The Two Children And The Witch
\item The Vain Queen
\end{tabitemize}

\\

    \bottomrule
  \end{tabular}
  \caption{The Fairy Tales included in the corpora. The Italian corpus includes several collectors. When not indicated, collected by Giuseppe Pitré. Otherwise, \label{editor:imbriani}1. Vittorio Imbriani\label{editor:comparetti}; 2. Domenico Comparetti\label{editor:gozenbach}; 3. Laura Gozenbach\label{editor:coronedi}; and 4. Carolina Coronedi-Berti.}
  \label{tab:texts}
\end{table*}

%% file: figures-tables/tab096-tokens-reduced.tex
\begin{table*}
\footnotesize
  \begin{tabular}{p{.15\textwidth}p{.55\textwidth}l}
  
    \toprule
Token & Synonyms & Value \\
    \midrule
dialogu & conversation & Universalism \\
equality & equality, equal & Universalism\\
free & free & Self-Direction \\
right & right, claim & Universalism \\
justic & justice, judge, trial, fairness, just & Universalism\\
peace & peace & Universalism\\
cooper & help, together & Benevolence\\
curios & curiosity, curious & Self-Direction\\
empathi & compassion, pity & Conformity\\
evid & evidence & Universalism\\
emancip & liberty & Self-Direction\\
generous & hospitality, goodness & Benevolence\\
honest & honest, confidence & Benevolence\\
smart & clever, cleverness, wise & Achievement\\
kind & kind, kindness, graciousness, gentleness & Conformity\\
know & know, able, knowledge & Achievement\\
brother & brother, sister, brotherly, sisterly & Benevolence\\
love & love, married, wife, husband, marriage, wedding & Benevolence\\
loyal & honor, faith & Benevolence\\
pieti & piety, pious, god, virgin, saint, angel, pray & Tradition\\
mother & mother, father, motherly, fatherly & Benevolence\\
punish & punish, punishment & Conformity\\
pure & pure, innocent, innocence & Tradition\\
correct & correct, reason, correctness & Universalism\\
reward & reward, prize, pay, treasure, jewels & Power\\
law & lawful, king, queen & Power\\
solidar & harmony, support & Benevolence\\
toler & acceptance, permissiveness & Universalism\\
truth & truth & Universalism\\

    \bottomrule
  \end{tabular}
  \caption{List of tokens mapped with the values proposed in the Theory of Basic Values~\citep{schwartz1992universals, lindeman_measuring_2005, Murteira_prep}.}
  \label{tab:tokens}
\end{table*}